\documentclass{article}
\usepackage{amssymb}
\usepackage{amsfonts}
\usepackage{lipsum}  % 用于生成示例文本
\usepackage{multicol}
\usepackage{multirow}
\usepackage{tabularx}
\usepackage{wrapfig}
\usepackage{booktabs}
\usepackage{makecell}
\usepackage{subfigure}
\usepackage{ijcai24}
\usepackage{times}
\usepackage{soul}
\usepackage{url}
\usepackage[hidelinks]{hyperref}
\usepackage[utf8]{inputenc}
\usepackage[small]{caption}
\usepackage{graphicx}
\usepackage{amsmath}
\usepackage{amsthm}
\usepackage{booktabs}
\usepackage{algorithm}
\usepackage{algorithmic}
\usepackage[switch]{lineno}

\title{Learning Attentional Mixture of LoRAs for Language Model Continual Learning}

\author{
Jialin Liu $\quad$ Jianhua Wu $\quad$ Jie Liu  $\quad$ Yutai Duan \\[1mm]
Nankai University, China \\
{\tt\small jliu@nankai.edu.cn} \\[-1mm]
}

\begin{document}

\maketitle

\begin{abstract}
% Real-world knowledge is continually evolving, requiring large language models (LLMs) to continuously learn and adapt to the latest developments. Addressing the issue of catastrophic forgetting has become a hot research topic. However, there are large differences in parameter space size and optimal solution distribution between LLMs and previous pre-trained language models, which may cause existing continuous learning methods to fail to achieve satisfactory results in LLMs. In this paper, we propose LoRA-Master - a continuous learning approach suitable for LLMs. It can dynamically combine all tasks in the continuous learning process, thereby effectively avoiding catastrophic forgetting. Specifically, LoRA-Master is trained while training the LoRA parameters of the new task, which is used to help the LoRA of the previous task adapt to the new scenario when a new task is added. In addition, we also add sparsity constraints so that LoRA-Master can effectively filter the output of multiple LoRAs. Experimental results on standard continuous learning benchmarks show that our proposed method outperforms existing methods. Finally, we illustrate the advantages of our method in LLM continuous learning scenarios by comparing LoRA-Master with previous continuous learning methods.

Fine-tuning large language models (LLMs) with Low-Rank adaption (LoRA) is widely acknowledged as an effective approach for continual learning for new tasks. However, it often suffers from catastrophic forgetting when dealing with multiple tasks sequentially. 
To this end, we propose Attentional Mixture of LoRAs (AM-LoRA), a continual learning approach tailored for LLMs. Specifically, AM-LoRA learns a sequence of LoRAs for a series of tasks to continually learn knowledge from different tasks. 
The key of our approach is that we devise an attention mechanism as a knowledge mixture module to adaptively integrate information from each LoRA. With the attention mechanism, AM-LoRA can efficiently leverage the distinctive contributions of each LoRA, while mitigating the risk of mutually negative interactions among them that may lead to catastrophic forgetting.
Moreover, we further introduce $L1$ norm in the learning process to make the attention vector more sparse. The sparse constraints can enable the model to lean towards selecting a few highly relevant LoRAs, rather than aggregating and weighting all LoRAs collectively, which can further reduce the impact stemming from mutual interference.
Experimental results on continual learning benchmarks indicate the superiority of our proposed method.
\end{abstract}

\section{Introduction}

Efficient parameter fine-tuning (PEFT) methods based on LoRA, as fundamental units for continual learning, are widely used to adapt large language models (LLMs) to downstream tasks from different domains ~\cite{trace}. 
However, it often suffers from catastrophic forgetting when dealing with multiple tasks sequentially. 

Several studies have explored the feasibility of LoRA-based approaches in continual learning for LLMs. Among them, O-LoRA ~\cite{olora} proposes to learn orthogonal low-rank adaptation for continual learning in language models. It is based on the assumption that the gradient subspace of previous tasks can be represented by LoRA parameters, which enables the model to progressively learn new tasks in an orthogonal subspace to mitigate catastrophic forgetting while learning new tasks.

\begin{figure}
  \centering
  \subfigure[Normal pre-trained model]{\includegraphics[width=0.45\linewidth]{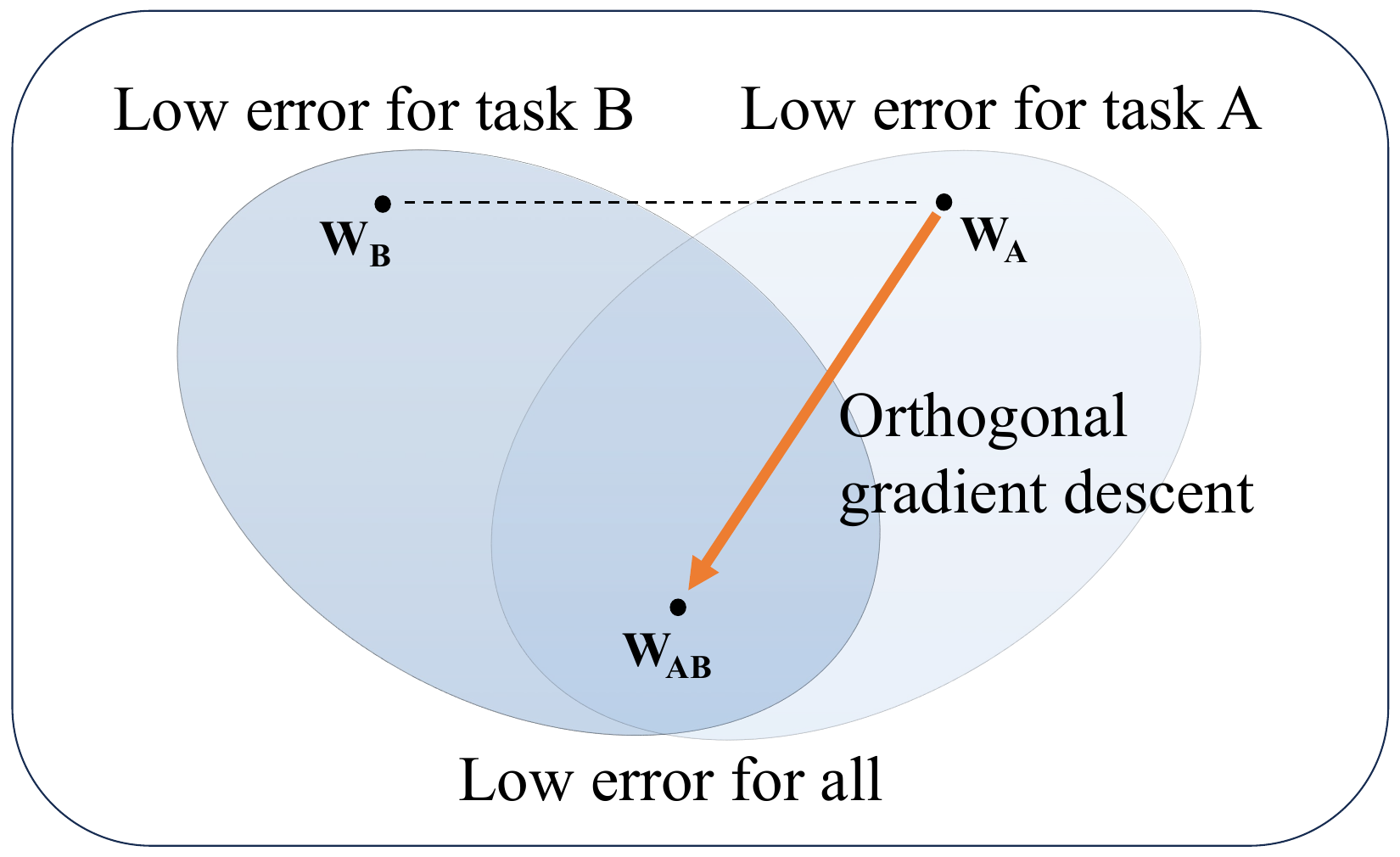}}
  \subfigure[Large language model]{\includegraphics[width=0.45\linewidth]{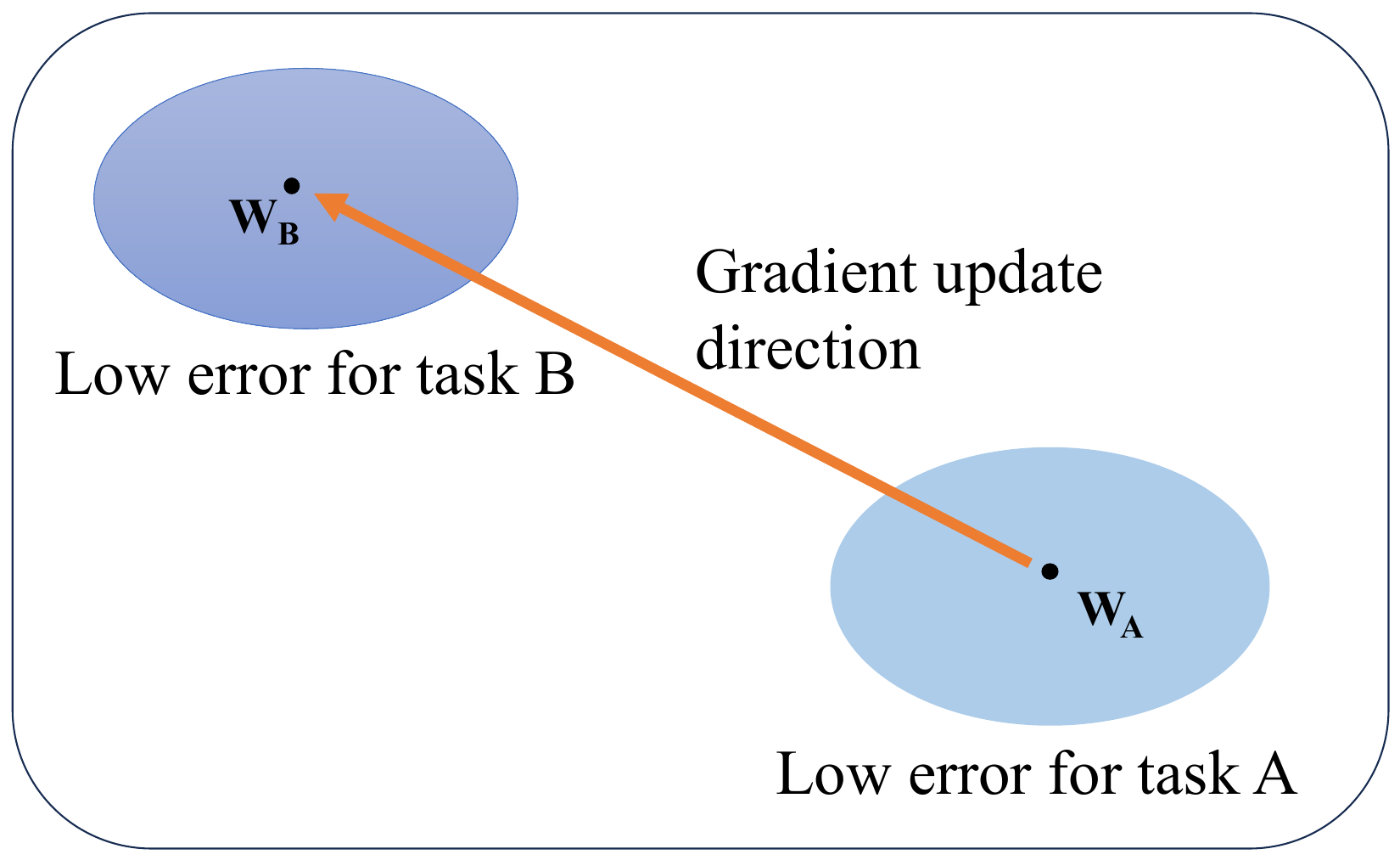}}
  \caption{Intuitive demonstration of the decentralized problem of optimal solutions. (a) is the distance relationship diagram between the possible optimal solutions of the two tasks in the normal pre-trained language model. There is a common optimal solution at the intersection of the two. The parameter space of the LLM (b) may be too large, causing the optimal solution areas of the two tasks to be too far apart, so that there is no common optimal solution.}
  \label{figure1}
\end{figure}

Despite the progress, existing approaches are still faced with important challenges.
Firstly, existing orthogonal gradient descent~\cite{farajtabar2020orthogonal} based approaches, e.g. O-LoRA, exhibits an issue of mismatch between tasks and parameters. From the perspective of parameter space, the assumption of orthogonal gradient descent is based on the existence of a common optimal solution for different tasks in the parameter space, as illustrated in Figure \ref{figure1}(a). However, even if the parameters converge to a common optimal solution, the model gradually deviates from the optimal parameters of the previous task, leading to the problem of mismatch between the current parameters and the parameters of the previous task. Especially in the huge parameter space of an LLM, the optimal parameters of multiple tasks may be extremely different from each other, and even no common optimal solution exists (Figure\ref{figure1}(b)). When learning a new task in this situation, the model parameters will match even less with the previous tasks, resulting in catastrophic forgetting. %We also conduct empirical analysis in our experiments.

In addition, existing approaches fail to dynamically adapt to the latest scenarios. During the training process, only the LoRA parameters of the new task are trainable, while other modules remain frozen. This prevents the LoRAs of previous tasks from adapting to the introduction of new tasks. Under such condition, simply adding up the outputs of each LoRA without precise knowledge selection may not only fail to prevent detrimental information from affecting the current task, but also result in heterogeneous conflicts between knowledge from new tasks and previous tasks.

To address these issues, we propose Attentional Mixture of LoRAs (AM-LoRA), which adaptively mixes different task-specific LoRAs' knowledge. Instead of focusing on subspace parameter update constraints, we design a novel framework that allows the model to effectively combine the capabilities of LoRAs learned squentially. 
Specifically, AM-LoRA consists of two components: Task-specific LoRA Matrix Sequences and an Attentional Selector. Task-specific LoRA Matrix Sequences are exploited to learn knowledge for various tasks, while the Attentional Selector is responsible for filtering and mixing knowledge from different LoRAs during the learning process to better solve the current task. This design enables the model to leverage appropriate knowledge for each task, avoiding issues of mismatch between tasks and parameters. AM-LoRA can also continuously adjust the contribution of each LoRA during the continual learning process, contributing to the model in dynamically adapting to new task scenarios. Additionally, we incorporate sparsity constraints into AM-LoRA, enabling precise selection of knowledge from different tasks. This mechanism not only reduces the impact of harmful information but also maximizes the utilization of beneficial knowledge to facilitate learning for new tasks, preventing heterogeneity conflicts among knowledge from different tasks.

The contribution of this paper can be summarized as follows:
\begin{itemize}
\item We propose a novel continual learning approach for LLMs, AM-LoRA, which can adaptively mix knowledge from a series of tasks conveyed by sequentially learned LoRAs based on an attention mechanism.

\item We introduce a sparse learning strategy for AM-LoRA, which results in a sparse attention vector and enables the model to lean towards selecting a few highly relevant LoRAs, rather than aggregating and weighting all LoRAs collectively, which can further reduce the impact stemming from mutual interference.
\item We conducted extensive experiments on real-world continual learning benchmarks. The experimental results show that our method achieves superior performance against existing SOTA methods.
\end{itemize}

\begin{figure*}
    \centering
    \includegraphics[width=1\linewidth]{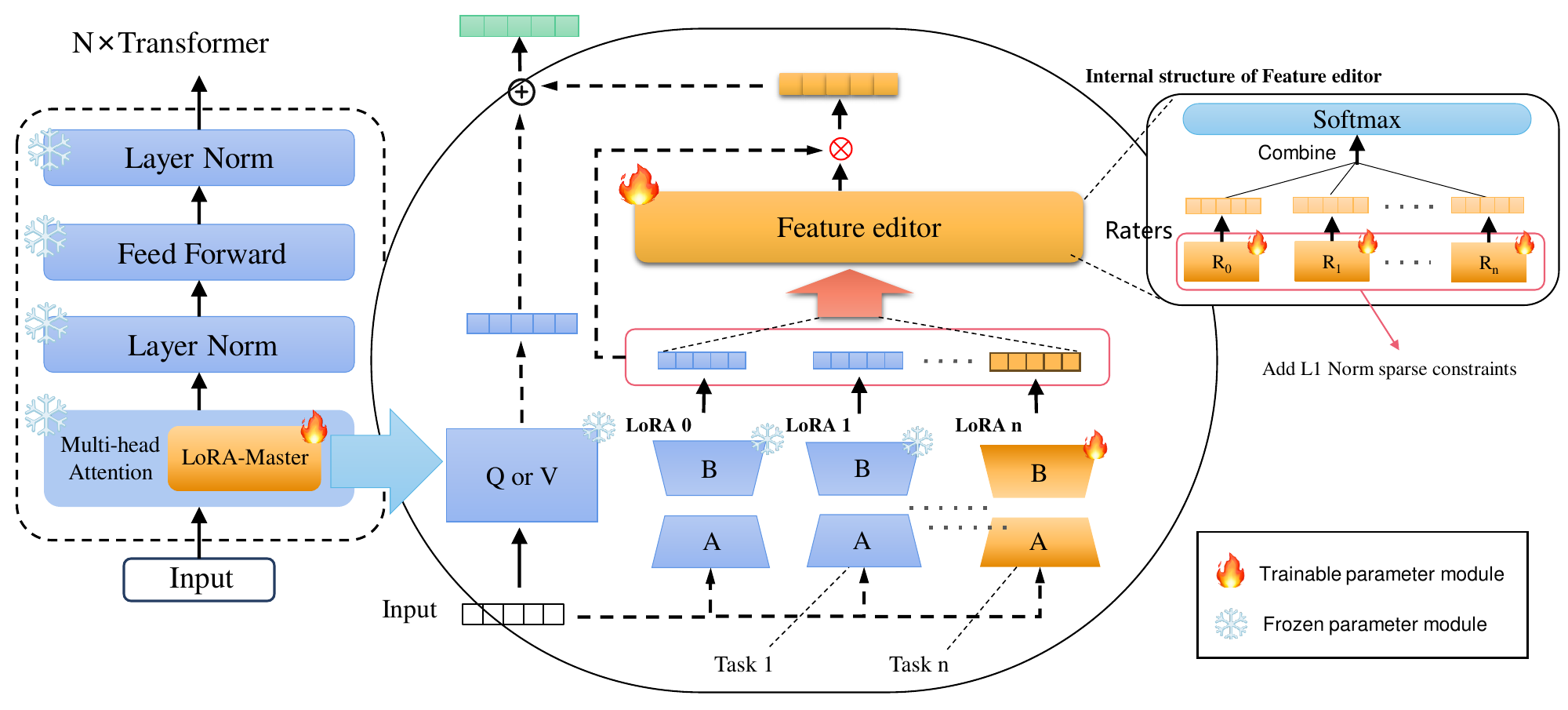}
    \caption{An overview of AM-LoRA. During the training of the current task, the weight of the pre-trained model and the LoRA parameters of the previous task are frozen to train the Attentional Selector and LoRA of the new task. In this process, Task-specific LoRA Matrix Sequences are mainly responsible for the learning of new task knowledge, while Attentional Selector focuses more on learning how to adaptively integrate information from each LoRA when a new task is added, making full use of the knowledge of previous tasks for efficient learning.}
    \label{Figure 2}
\end{figure*}

\vspace{-2.5pt}
\section{Related Work}
\vspace{-2.5pt}

In this section, we first introduce Parameter Efficient Fine-Tuning (PEFT) techniques for language models, followed by an introduction to existing continuous learning methods. These two types of work are closely related to our approach.

\subsection{Parameter Efficient Fine-Tuning}

% Large language models (LLMs), such as GPT-3~\cite{gpt3}, PaLM~\cite{palm}, and LLaMA~\cite{llama}, have hundreds of billions or even more parameters, which enable them to understand natural language better and generate high-quality text based on a given context (e.g., prompt). However, it is necessary to retrain LLMs to adapt to the continuous updating of real-world knowledge~\cite{LLMsurvey,LLMsurvey2}. So continual learning is crucial for existing LLMs~\cite{trace}. 

Parameter Efficient Fine-Tuning (PEFT) aims to optimize the fine-tuning process to minimize the computational cost and time required for LLMs to adapt to new tasks or new knowledge~\cite{adapter,zhu2021counter,DBLP:conf/iclr/HeZMBN22,karimi2021compacter}. The more commonly used PEFT methods include Adapter, Prefix Tuning, P-Tuning, and LoRA. among them, LoRA is gradually becoming a common paradigm for instruction fine-tuning in LLMs due to its superior performance.

Some existing work focuses on using LoRA for continual learning based on LLMs. However, how to solve the catastrophic forgetting problem is a core challenge. The purpose of our method is to address this issue.

\subsection{Continual Learning}

Continual learning(CL) refers to a process in which a model learns a sequence of tasks or data streams continuously while preserving the knowledge and capabilities acquired from previous tasks when learning new tasks~\cite{chen2018lifelong}. Existing CL methods can be categorized into three types. 

\textit{Rehearsal-based methods} score and replay a small part of training samples from previous tasks or learn to generate pseudo samples of previous tasks ~\cite{rebuffi2017icarl,replay_lopez,shin2017continual,DBLP:conf/iclr/KemkerK18}. 
% The drawback of this type of approach is that samples from previous tasks need to be replayed during training for subsequent tasks, which raises data privacy concerns.
\textit{Regularization-based methods} add additional loss to consolidate the parameters learned from previous tasks in learning a new task~\cite{regular_kirk,regular_zenke,he2018overcoming,ebrahimi2019uncertainty}. 
% This method of limiting the direction of model gradient update still cannot avoid catastrophic forgetting because it will also gradually move away from the optimal solution of the previous task.
\textit{Architecture-based methods} learn different set of parameters for a separate task~\cite{rusu2016progressive,mallya2018packnet,mallya2018piggyback}. 
% Progressive Prompts~\cite{razdaibiedina2023progressive} get individual prompts for each incoming task and connect them in order with the previously learned prompts. 
% However, this approach essentially recognizes task ID, which limits their ability to generalize to unknown tasks. 

Due to the high training cost and the challenge of obtaining high quality data, it is crucial to use the PEFT method for continuous learning of LLMs, which also belongs to the parameter isolation method.
There are a number of current studies that have explored the effectiveness of using the PEFT approach to continuous learning for LLMs, such as O-LoRA~\cite{olora}, which innovatively adopts LoRA for continual learning in LLM. However, O-LoRA simply adds all LoRA output features, which causes aggravated information loss, so it also has the drawbacks such as catastrophic forgetting.

As discussed earlier, our approach is more suitable for LLM continual learning scenarios than existing approaches, allowing LLM to learn new tasks efficiently while avoiding catastrophic forgetting.

\vspace{-2.5pt}
\section{Method}
\vspace{-2.5pt}

In this section, we propose AM-LoRA to adaptively integrate knowledge from various task-specific LoRAs via an attention mechanism. We first briefly introduce LoRA, which is the basis of our approach. Then we present a brief overview of our methodological framework. The following two subsections detail the two core parts of our method. Finally, we will discuss the key differences between AM-LoRA and O-LoRA and the superiority of our approach.
\vspace{-2.5pt}
\subsection{Preliminary}
\vspace{-2.5pt}
Low-Rank Adaptation (LoRA)  ~\cite{lora} is a low-resource fine-tuning method for Large Language Models (LLMs) based on the low-rank assumption. It is composed of two low-rank matrices, \(B \in \mathbb{R}^{d \times r}\) and \(A \in \mathbb{R}^{r \times k}\), where \(r \ll \min(d, k)\). Here, \(A\) is responsible for dimensional reduction matrix, and \(B\) is for dimensional expansion matrix. During fine-tuning, the parameters of the Large Language Model are frozen, and only \(B\) and \(A\) are trained.

Due to the small value of \(r\), fine-tuning becomes computationally efficient. The update of LLM's parameters can be expressed as:
\begin{equation}
W = W_0 + BA, \tag{1}
\end{equation}
where \(W_0\) and \(W \in \mathbb{R}^{d \times k}\) represent the parameters before and after the update of \(\mathcal{M}\), respectively. Typically, the matrix \(BA\) is denoted as \(\Delta W\).

\vspace{-2.5pt}
\subsection{Framework}
\vspace{-2.5pt}

As shown in Figure \ref{Figure 2}, AM-LoRA consists of two parts: a Task-specific LoRA Matrix Sequence and an Attentional Selector responsible for combining all LoRA capabilities. Among them, the LoRA matrix sequence is mainly responsible for learning new task knowledge. The Attentional Selector is more focused on learning how to filter useful knowledge in LoRAs to better handle new tasks. We will introduce these two parts in detail in the next two subsections. The input data passes through LoRA for different tasks at the same time, and then features with corresponding task knowledge are generated. All features are then sparsely selected and attentional mixed by the Attentional Selector, ultimately producing high-quality features that fully combine all task knowledge.
\vspace{-2.5pt}
\subsection{Incremental Learning of Task-specific LoRAs}
\vspace{-2.5pt}
Due to the fact that the original method requires adjusting all parameters of LoRA when learning each task, it makes it easy for a single LoRA to disrupt the knowledge of previous tasks when sequentially learning multiple tasks. Therefore, we adopt an incremental learning approach to learn an independent LoRA for each task, and the LoRAs of all tasks together form a task-specific LoRA sequence.

Specifically, as shown at the bottom of Figure \ref{Figure 2}, given a task sequence \(T = \{t_1, t_2, ..., t_N\}\) and its corresponding training dataset \(D = \{d_1, d_2, ..., d_N\}\), the LLM \(M\) continuously fine-tunes to obtain a Task-specific LoRA Matrix Sequence \(\Delta W = \{\Delta w_1, \Delta w_2, ..., \Delta w_n\}\). We illustrate the learning process for the \(n\)th task. We freeze all parameters of the pre-trained model \(M\) and the LoRA matrices \(\Delta w_0\) to \(\Delta w_{n-1}\) from previous tasks. Only the new task's LoRA \(\Delta w_n\) is involved in training. Under this setting, the forward process of the LoRA training for the \(n\)th task can be expressed as follows:

\begin{equation}
W = W_{0} + \sum_{i=1}^{n-1} \Delta w_{i}, \tag{2}
\end{equation}

\begin{equation}
\begin{aligned}
h &= Wx + \Delta W_{n}x \\
&= Wx + B_{n}A_{n}x,
\end{aligned}
\tag{3}
\end{equation}
where \(x\in \mathbb{R}^{d_{in}}\) is the input vector, and \(h\in \mathbb{R}^ {d_{out}}\) is the output vector. \(W_{0}\) is the key and value projection matrix of transformer blocks, and \(\Delta w_{1}, \Delta w_{2}, ..., \Delta w_{n-1}\) are the LoRA matrices for the previous \(n-1\) tasks, all of which are frozen. However, merely freezing the LoRA parameters of previous tasks is not sufficient to address catastrophic forgetting. Simply summing all LoRA features during inference will lose information from past tasks, resulting in performance degradation on past tasks. To address this issue, we propose AM-LoRA, which will be detailed in the next subsection.

\vspace{-2.5pt}
\subsection{Attentional Selector}
\vspace{-2.5pt}
Attentional Selector is a core part of our approach, based on an attention mechanism we designed to efficiently integrate knowledge in task-specific LoRAs. Specifically, the Attentional Selector is added and trained together with the new task's LoRA matrix $\Delta w_n$. As shown on the right side of Figure \ref{Figure 2}, input the features of each task-specific LoRA to the Attentional Selector, and then perform non-linear transformation through the corresponding dense layer, which is denoted as \(W_{\text{ri}} \in \mathbb{R}^{k \times 1}\) (where \(i=0, \ldots, n\)). Next, the transformed LoRA features are combined together and passed through the softmax function. Then the attention scores of each task can be obtained in this state:

\begin{equation}
\begin{aligned}
g_i &= \text{Softmax}(R_i(\Delta w_i x)) \\
&= \text{Softmax}(W_{ri}^T(\Delta w_i x)),
\end{aligned}
\tag{4}
\end{equation}
where $R_i(x)$ ($i=0,...,n$) represents the corresponding dense layer for each LoRA, and $x \in \mathbb{R}^{b \times k}$ is the input vector. Then the forward process after adding Attentional Selector can be expressed as:

\begin{equation}
h = W_0 + \sum_{i=0}^{n} g_i \cdot (\Delta w_i x). \tag{5}
\end{equation}

Inspired by the residual connection in ResNet~\cite{he2016deep}, we add a zero LoRA matrix $\Delta w_0$, allowing the model to choose not to utilize the knowledge from previous tasks while learning new tasks. Additionally, it serves the purpose of assigning weights when training the LoRA for the first task.
\vspace{-2.5pt}
\subsection{Loss Function with Sparsity Constraint}
\vspace{-2.5pt}
We have observed that the model may retain features that are irrelevant or harmful to the current task during the learning process, leading to heterogeneous conflicts in the knowledge of different tasks. This will have a negative impact on the model's generalization performance and learning effectiveness. To solve these problems, we introduce the L1 norm to make the attention vector sparser. This can promote the model to perform LoRA selection more accurately and reduce mutual interference between different tasks. We just add sparsity constraints on the dense layer of Attentional Selector, which will not affect the learning effect of LoRA sequences. The training loss function of the \(n\)th task is as follows:

\begin{equation}
L_{total} = L_{task} + \lambda \sum_{i=0}^{n} \lVert W_{R_i} \rVert_1, \tag{6}
\end{equation}
where $L_{\text{task}}$ represents the loss for the current task, $W_{R_i}$ ($i=0, \ldots, n$) denotes the coefficient matrices of n+1 dense layers, and $\lambda$ is the weight for the L1 norm loss. As the learning tasks gradually increase, this design will improve the generalization ability of the model. This enables it to better mitigate catastrophic forgetting in the context of longer task sequences.

\begin{table*}
    \centering
    \begin{tabularx}{\linewidth}{p{2cm}XXXXXXXX}
        \toprule
         & \multicolumn{3}{c}{\textbf{Standard CL benchmarks}} & & \multicolumn{3}{
         c}{\textbf{Large Number of Tasks}}\\[5pt]
        
         & \textbf{Order1} & \textbf{Order2} & \textbf{Order3} & \textbf{Avg} & \textbf{Order4} & \textbf{Order5} & \textbf{Order6} & \textbf{Avg}\\
        \midrule
        SeqFT & 18.9 & 24.9 & 41.7 & 28.5 & 7.4 & 7.4 & 7.5 & 7.4\\
        SinLoRA & 44.6 & 32.7 & 53.7 & 43.7 & 2.3 & 0.6 & 1.9 & 1.6\\
        IncLoRA & 66 & 64.9 & 68.3 & 66.4 & 63.3 & 58.5 & 61.7 & 61.2\\
        Replay & 55.2 & 56.9 & 61.3 & 57.8 & 55 & 54.6 & 53.1 & 54.2\\
        EWC & 48.7 & 47.7 & 54.5 & 50.3 & 45.3 & 44.5 & 45.6 & 45.1\\
        L2P & 60.3 & 61.7 & 61.1 & 60.7 & 57.5 & 53.8 & 56.9 & 56.1\\
        LFPT5 & 67.6 & 72.6 & \textbf{77.9} & 72.7 & 70.4 & 68.2 & 69.1 & 69.2\\
        O-LoRA & 75.4 & 75.7 & 76.3 & 75.8 & 72.3 & 64.8 & 71.6 & 69.6\\
        \midrule
        \textbf{Ours} & \textbf{78.1} & \textbf{79.8} & 76.2 & \textbf{78.0} & \textbf{72.7} & \textbf{73.3} & \textbf{71.8} & \textbf{72.6}\\
        \midrule
        ProgPrompt & 75.2 & 75 & 75.1 & 75.1 & 78 & 77.7 & 77.9 & 77.9 \\
        PerTaskFT & 70 & 70 & 70 & 70 & 78.1 & 78.1 & 78.1 & 78.1 \\
        MTL & 80 & 80 & 80 & 80 & 76.5 & 76.5 & 76.5 & 76.5 \\
        \bottomrule
    \end{tabularx}
    \caption{Summary of results on Standard CL benchmarks and Large Number of Tasks benchmarks using T5-large models with AM-LoRA. Report the average accuracy of all tasks after training for the last task. All results were averaged over 3 runs.}
    \label{Table 1}
\end{table*}
\vspace{-2.5pt}
\subsection{AM-LoRA vs. O-LoRA}
\vspace{-2.5pt}
% AM-LoRA is a rehearsal-free method~\cite{de2019episodic,huang2021continual} that avoids catastrophic forgetting by effectively integrate information from different LoRAs through the attention mechanism. There is no need to store any data from past tasks in the training of subsequent tasks, so our approach does not raise data privacy issues.
% Moreover, our method does not depend on the recognition of task ID~\cite{razdaibiedina2023progressive,wang2022learning}, which is harmful to the generalization of the model. In the process of training a series of tasks, we adopt the method of instruction tuning. The strategy of instruction tuning enables our model to better understand the nature of tasks, adapt flexibly to new tasks, and not be bound by specific task ids.

O-LoRA~\cite{olora} is also essentially a regularization-based approach by learning in a subspace orthogonal to the LoRA subspace associated with the previous task. This approach limits the model's ability to capture inter-task heterogeneity, which in turn affects the learning of subsequent tasks. Furthermore, even restricting the new task LoRA parameters to be updated in the orthogonal space of the previous gradient space gradually moves the parameters away from the optimal solution of the past task, leading to catastrophic forgetting. For a detailed mathematical discussion on this point, we refer to the Appendix.

Instead of adding too many constraints to the parameter update of the task, AM-LoRA is filtered by filtering the features from the LoRA output of the previous task. This design allows for more flexible adaptation to the features of different tasks and improves the effectiveness of learning for new tasks. Moreover, by selectively retaining knowledge that is useful for the current task, AM-LoRA avoids heterogeneous conflicts between individual task features, thus reducing the impact of catastrophic forgetting.

\vspace{-2.5pt}
\section{Experiments}
\vspace{-2.5pt}

In this section, we evaluate the performance of AM-LoRA through a series of continual learning experiments, which are summarized to answer the following research questions:
\begin{itemize}
\item \textbf{RQ1}: How does AM-LoRA perform on the continual learning benchmark compared to various baselines?

\item \textbf{RQ2}: Does the decentralized problem of optimal solutions harm the performance of continual learning methods for LLMs?

\item \textbf{RQ3}: How does AM-LoRA avoid catastrophic forgetting during continual learning?

\item \textbf{RQ4}: How do designed different sub-modules contribute to the model performance?
\end{itemize}
\vspace{-2.5pt}
\subsection{Datasets}
\vspace{-2.5pt}
% \subsubsection{Standard CL benchmarks}
We first conducted experiments on \textit{the Standard CL benchmarks}~\cite{zhang2015character}, which includes five text classification datasets: AG News, Amazon Reviews, Yelp reviews, DBpedia and Yahoo Answers. We adopt the CL settings of the T5 model ~\cite{raffel2020exploring}, following O-LoRA~\cite{olora}, and explore three different orders of benchmarks. We provide task details in Appendix A.1 and our experiments in Appendix A.2.

% \subsubsection{Large number of tasks}

Following \textit{the Large number of tasks} proposed by ~\cite{razdaibiedina2023progressive}, longer task sequences pose greater challenges to our approach, which is evaluated by experiments on a continual learning benchmark on 15 datasets. This includes five tasks in the Standard CL benchmarks, four tasks in the GLUE benchmark (MNLI, QQP, RTE, SST2)~\cite{wang2018glue}, five tasks in the SuperGLUE benchmark (WiC, CB, COPA , MultiRC, BoolQ) ~\cite{wang2019superglue} and the IMDB movie review dataset~\cite{maas2011learning}. Following the previous approach~\cite{olora}, we select 1000 random samples for training on each task and retain 500 samples for each category for validation.
\vspace{-2.5pt}
\subsection{Comparison Methods}
\vspace{-2.5pt}
To demonstrate the effectiveness of AM-LoRA, we compare our model with the following methods:

\begin{itemize}
    \item \textbf{Methods based on full fine-tuning}: \textit{SeqFT}~\cite{de2019episodic} fine-tunes all parameters of a pre-trained model across a series of tasks. No regularization or replaying of samples from previous tasks is required. \textit{PerTaskFT} trains a separate model for each task.
    \item \textbf{Basic methods based on LoRA}: \textit{SinLoRA} only uses one LoRA to train on a series of tasks without adding any regularization constraints and past sample replay. As an improvement of SinLoRA, \textit{IncLoRA} adds a LoRA module when training new tasks. Each task has its own LoRA and no regularization or sample replay is added.
    \item \textbf{Continual learning baseline methods}: \textit{Replay} leverages memory buffering to fine-tune the entire model and replay samples from previous tasks during learning of new tasks. \textit{EWC} ~\cite{regular_kirk} fine-tunes the entire model using a regularization loss, which is used to suppress parameter updates that may corrupt knowledge of previous tasks. \textit{L2P}~\cite{wang2022learning}, \textit{LFPT5}~\cite{qin2021lfpt5} and \textit{ProgPrompt}~\cite{razdaibiedina2023progressive} all train specific prompts for all tasks during the training phase. In the inference phase, the trained hints are used as replay samples or markers to determine task IDs. \textit{O-LoRA}~\cite{olora} enables the model to incrementally learn new tasks in orthogonal subspaces by freezing the LoRA parameters of past tasks. It is the current SOTA method of continual learning. 
    \item \textbf{MTL}: Training all tasks in a model in a multi-task learning manner is often considered the upper limit of continual learning.
\end{itemize}
\vspace{-2.5pt}
\subsection{Experimental Settings}
\vspace{-2.5pt}
In order to compare with recent continual learning methods, we performed experiments on the widely used pre-trained T5-Large model~\cite{raffel2020exploring}. In addition, in order to check the effect of our method in LLMs, We performed experiments using the LLaMA2-7B model~\cite{touvron2023llama}. For LoRA settings, we adopt the matrix rank $r$ of 8 and the scale factor $\alpha$ is set to 32. All of our experiments on both models were conducted on a machine equipped with eight NVIDIA GeForce RTX 3090 and the implementation using DeepSpeed repository. All experimental results are the average of the results of the three runs. We report more experimental details in Appendix A.2.

\begin{table}
    \centering
    \begin{tabularx}{\linewidth}{XXXXX}
        \toprule
        & \textbf{Order1} & \textbf{Order2} & \textbf{Order3} & \textbf{Avg} \\[3pt]
        \midrule
        IncLoRA & 74.5 & 71.2 & 72.9 & 72.8 \\[3pt]
        O-LoRA & 76.8 & 75.7 & 75.7 & 76.1 \\[3pt]
        \textbf{Ours} & \textbf{78.8} & \textbf{78.5} & \textbf{78.4} & \textbf{78.6} \\[1pt]
        \midrule
        MTL & \multicolumn{4}{c}{77.1} \\
        \bottomrule
    \end{tabularx}
    \vspace{5pt}
    \caption{Summary of results of AM-LoRA combined with LLaMA2-7B model on Standard CL benchmarks. And a comparison with the same LoRA-based baseline methods: IncLoRA, O-LoRA, and multi-task learning "MTL" using the same model.}
    \label{Table 2}
\end{table}

\subsection{AM-LoRA Performance on continual learning Benchmarks (RQ1)}

% \subsubsection{Performance comparison of AM-LoRA with other continual learning methods under two CL benchmarks}

Table \ref{Table 1} shows the performance comparison between AM-LoRA and the baseline continuous learning approach on the standard CL benchmark and a large number of task benchmarks. Based on previous work, we report the results of 3 runs on the T5-Large model using different task sequences.

\subsubsection{Results on Standard CL benchmarks} 
In experiments with Standard CL benchmarks for T5-Large model, our method significantly outperforms all continual learning baseline methods in order1, order2 and average results, and is competitive with SOTA in order3. This indicates that AM-LoRA has superior performance compared to previous continuous learning methods in the same experimental settings.
% In the previous analysis, we mentioned that the optimal solution common to multiple tasks may exist in normal pre-trained model, and the essence of multi-task learning is to find the optimal solution common to multiple tasks, and simultaneous learning will not produce catastrophic forgetting, so it is generally considered that multi-task learning is the upper limit of continual learning methods.
Besides, it can be seen that the effect of AM-LoRA is very close to that of multi-task learning, and is obviously better than that of previous continual learning methods, especially in order2, which further proves that our method is more effective in solving the problem of catastrophic forgetting. 

\subsubsection{Results on Large Number of Tasks} 
Our method also shows competitive performance in Large number of tasks, surpassing most baseline methods. However, there are still some gaps compared to PerTaskFT and MTL methods. It can be seen that longer task sequences are still a challenge for continual learning methods. In addition, ProgPrompt~\cite{razdaibiedina2023progressive} has shown excellent performance in longer sequence tasks. It essentially learns a separate prompt for each task and then performs task identification during inference. However, as mentioned in previous work, this method will reduce the generalization of the model and is not suitable for LLM. We believe that the goal of continual learning is to allow the model to continuously improve its capabilities as it continues to learn tasks, not just to identify tasks.
\vspace{-2.5pt}
\subsection{Empirical research on decentralized problem of optimal solutions (RQ2)}
\vspace{-2.5pt}
Table \ref{Table 2} shows the performance comparison between AM-LoRA and each baseline method on LLaMA2-7B. The O-LoRA and IncLoRA methods did not perform better on LLaMA2-7B than on T5-Large, even though LLaMA2-7B is a larger and more powerful LLM. Not only in continuous learning methods, multi-task learning (MTL) is not as effective in LLaMA2-7B as it is on T5-Large. The experimental results reinforce our previous analysis that the parameter space of LLM is much larger compared to a normal pre-trained model, and thus the distance between the optimal solutions of multiple tasks may be too far, resulting in no common optimal solution, and thus making it difficult to achieve the performance of a smaller model. 

AM-LoRA filters and efficiently combines the outputs of multiple LoRAs to facilitate the learning of a new task without destroying the capability of each task while ensuring that the knowledge of previous tasks is not destroyed. leverages the knowledge from previous tasks to facilitate the learning of new tasks.
Thanks to this advantage, our method outperforms MTL in LLMs. In addition, it should be clarified that MTL is the upper limit of the O-LoRA method, but not the upper limit of our method. The upper bound of our method is to train a model for each task separately, and that pedestal model is sufficiently powerful such that the method can completely avoid catastrophic forgetting.
% \subsection{Analysis}
\vspace{-2.5pt}
\subsection{Effectiveness of AM-LoRA in LLM continual learning (RQ3)}
\vspace{-2.5pt}
\begin{figure}[htbp]
    \centering
    \includegraphics[width=0.9\linewidth]{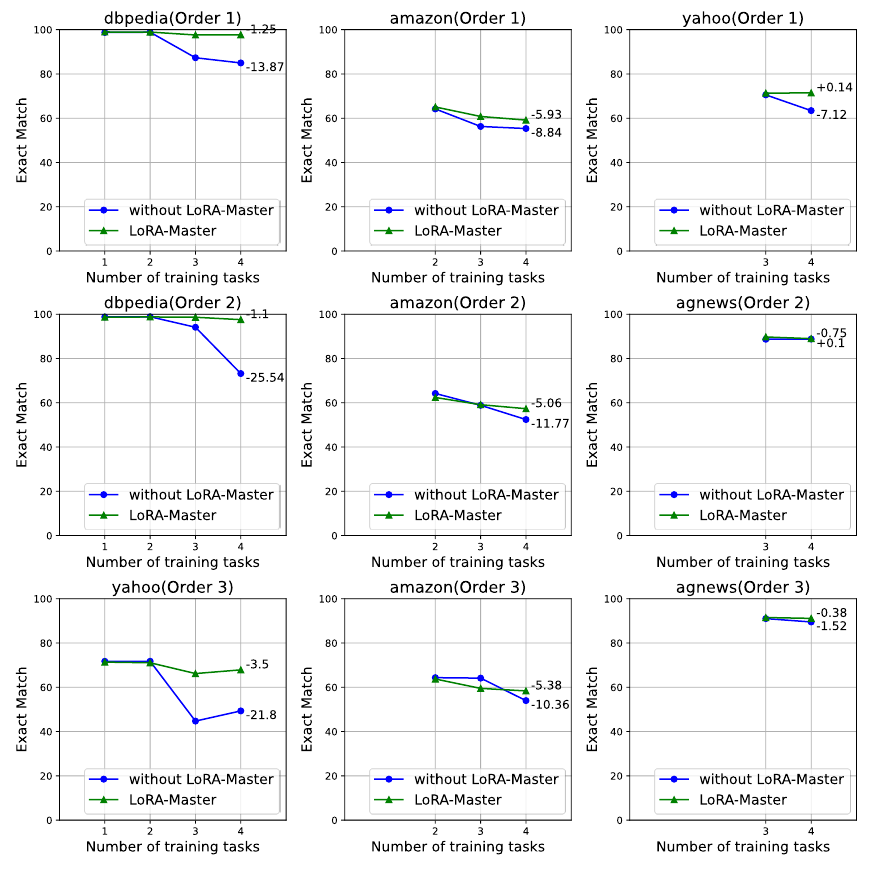}
    \caption{Comparison of LLaMA2-7B models with and without AM-LoRA on Standard CL benchmarks. We report the change of metrics separately for each task as the training task increases, and since the fourth task is the last one where no trend change can be observed, we only show the situation for the first three tasks in each order.}
    \label{Figure 3}
\end{figure}

To demonstrate that AM-LoRA can effectively mitigate catastrophic forgetting, we compared with models that are not equipped with AM-LoRA (learning a separate LoRA for each task, freezing the previous LoRA parameters when learning the LoRA for a new task), the performance of each task was recorded over time.

As shown in Figure \ref{Figure 3}, we can observe that the method of freezing the previous LoRA while learning the new task can maintain the ability of the past task to some extent. After learning two tasks, there is a small drop in the performance of the previous task. But after training four tasks, there is a large drop in the performance of the first task. The strategy of freezing the parameters of the visible basis is not enough to solve the catastrophic forgetting problem. When model with AM-LoRA learns multiple tasks, the performance of the previous tasks can still be maintained at a high level, and the ability to maintain the knowledge of the previous tasks has been greatly improved compared with the previous methods. It can be seen that AM-LoRA can effectively alleviate catastrophic forgetting by screening and combining the output features of each LoRA.

% Amazon and Yahoo are two difficult text categorization datasets, which is reflected in all the experimental results of Baselines. In addition, AM-LoRA performs better than previous methods on these two datasets. In this paper, considering only from the perspective of forgetting, the performance degradation of AM-LoRA on the Yahoo dataset in Order1 and Order3 is much smaller than that of the model without the addition of AM-LoRA, which proves that AM-LoRA can effectively avoid catastrophic forgetting.

\begin{figure}[htbp]
  \centering
  \includegraphics[width=0.5\linewidth]{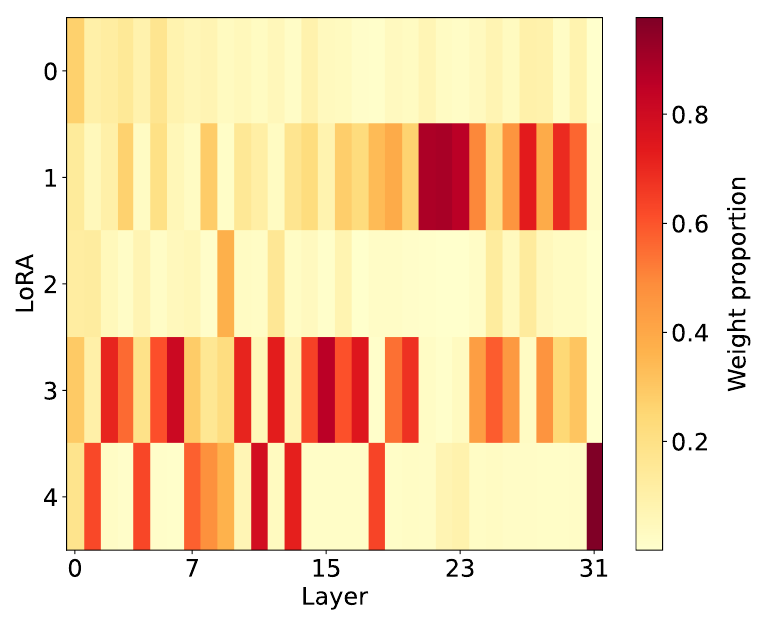}
  \includegraphics[width=0.5\linewidth]{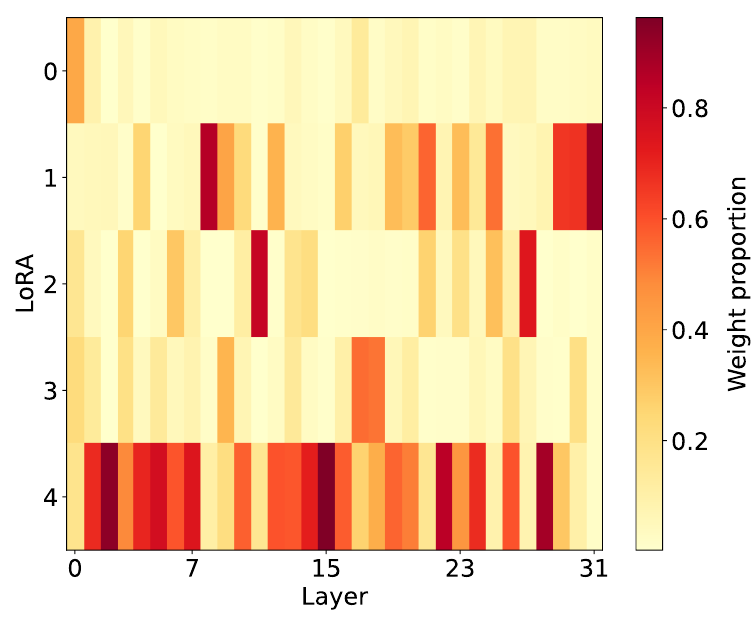}
  \caption{Attention score distribution of each LoRA in query matrix and value matrix (LLaMA2-7B). (a) is the weight distribution diagram of AM-LoRA with query matrix bypass. It can be observed that it mainly uses the knowledge in LoRA1, LoRA3, and LoRA4. The bypass of the value matrix (b) is more inclined to utilize the knowledge in the LoRA of this task (LoRA4).}
  \label{Figure 4}
\end{figure}

\begin{figure}[htbp]
  \centering
  \includegraphics[width=\linewidth]{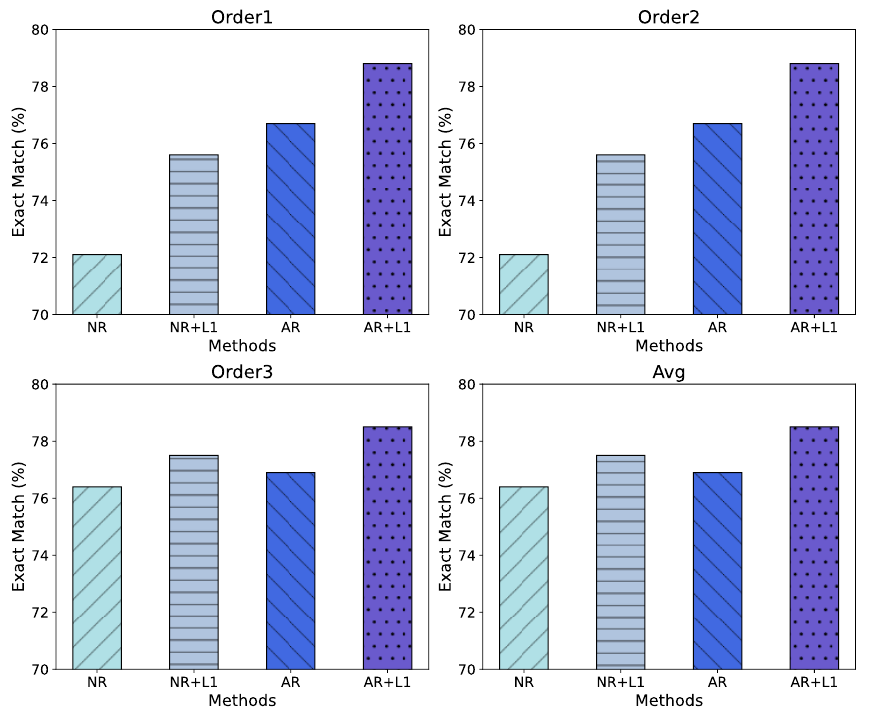}
  \caption{Comparison of the effectiveness of different AM-LoRA schemes on the Standard CL benchmarks. Among them, NR means that only the new task's dense layer participates in the training, AR means that the dense layers of all tasks participate in the training, and L1 means that sparse constraints are added to dense layers.}
  \label{Figure 5}
\end{figure}

As shown in Figure \ref{Figure 4}, we selected a piece of data in dbpedia to observe the attention score distribution of AM-LoRA to each task LoRA as the number of layers continues to deepen. It can be observed that the AM-LoRA bypassing the query matrix is more inclined to use the knowledge of LoRA of other tasks, while the AM-LoRA of the value matrix is more inclined to use the knowledge of LoRA of its own task. This further proves that our method better utilizes all the knowledge of LoRA to solve the catastrophic forgetting problem.

% \begin{wrapfigure}{l}{0.45\linewidth}
%   \centering
%   \subfigure[Query matrix]{\includegraphics[width=0.45\linewidth]{figure4-1-cropped.pdf}}
%   \subfigure[Value matrix]{\includegraphics[width=0.45\linewidth]{figure4-2-cropped.pdf}}
%   \caption{Attention score distribution of each LoRA in query matrix and value matrix(LLaMA2-7B). (a) is the weight distribution diagram of AM-LoRA with query matrix bypass. It can be observed that it mainly uses the knowledge in LoRA1, LoRA3, and LoRA4. The bypass of the value matrix(b) is more inclined to utilize the knowledge in the LoRA of this task(LoRA4).}
%   \label{Figure 4}
% \end{wrapfigure}

% \begin{wrapfigure}{r}{0.45\linewidth}
%     \centering
%     \includegraphics[width=\linewidth]{figure5-cropped.pdf}
%     \caption{Comparison of the effectiveness of different AM-LoRA schemes on the Standard CL benchmarks. Among them, NR means that only the new task's dense layer participates in the training, AR means that the dense layers of all tasks participate in the training, and L1 means that sparse constraints are added to dense layers.}
%     \label{Figure 5}
% \end{wrapfigure}

\vspace{-2.5pt}
\subsection{Ablation Studies (RQ4)}
\vspace{-2.5pt}
In this section, we tried four different AM-LoRA schemes, where the changes were the number of trainable dense layers in the Attentional Selector and whether to add L1 norm. As shown in figure \ref{Figure 5}, when only the dense layer corresponding to the new task participates in training, the previous LoRA dense layer is frozen. This can have the benefit of reducing the number of training parameters, but it can be observed that the effectiveness of this solution is not ideal. The role of the dense layer in AM-LoRA is to help past tasks adapt to the new state when learning the LoRA of a new task, so freezing the dense layer of the previous task would lead to a decrease in capacity. 

% After it was found that increase training a group of LoRA + dense layer can only increase the number and model of the overall number of 0.03\%, so this kind of lightweight structure in the task a limited number of cases is acceptable, because it brings the certain effect.

Furthermore, we can observe that adding sparse constraints in dense layers can effectively shield harmful features and enable the model to accurately select appropriate knowledge. This design can also improve the generalization of the model and better reflect its advantages when learning longer task sequences.

\vspace{-2.5pt}
\section{Conclusion}
\vspace{-2.5pt}

In this paper, we propose AM-LoRA, which can effectively mix the knowledge of each specific task LoRA based on attention to deal with catastrophic forgetting problem. Moreover, we introduce sparsity constraints to the learning of AM-LoRA, enabling it to accurately select appropriate knowledge in LoRAs. While mitigating catastrophic forgetting, our method can also allow the model to more rationally utilize knowledge from previous tasks. Extensive experiments on continuous learning benchmarks demonstate that our proposed AM-LoRA achieves superior performance against existing SOTA continuous learning methods.

\vspace{-2.5pt}
\section{Limitations}
\vspace{-2.5pt}

Although AM-LoRA outperforms previous LLM continuous learning methods in continuous learning benchmarks, there are some limitations in real-world applications. When the number of tasks is increasing, the increase in the number of LoRAs leads to an increase in the training cost as well. How to achieve continuous learning for a large number of tasks using a limited number of LoRAs in real-world scenarios is a challenge and is the work we focus on in the next phase.

\clearpage
\bibliographystyle{named}
\bibliography{ijcai24}

\begin{thebibliography}{}

\bibitem[\protect\citeauthoryear{Chen and Liu}{2018}]{chen2018lifelong}
Zhiyuan Chen and Bing Liu.
\newblock {\em Lifelong machine learning}, volume~1.
\newblock Springer, 2018.

\bibitem[\protect\citeauthoryear{de Masson~D'Autume \bgroup \em et al.\egroup }{2019}]{de2019episodic}
Cyprien de~Masson~D'Autume, Sebastian Ruder, Lingpeng Kong, and Dani Yogatama.
\newblock Episodic memory in lifelong language learning.
\newblock {\em Advances in Neural Information Processing Systems}, 32, 2019.

\bibitem[\protect\citeauthoryear{Ebrahimi \bgroup \em et al.\egroup }{2019}]{ebrahimi2019uncertainty}
Sayna Ebrahimi, Mohamed Elhoseiny, Trevor Darrell, and Marcus Rohrbach.
\newblock Uncertainty-guided continual learning with bayesian neural networks.
\newblock {\em arXiv preprint arXiv:1906.02425}, 2019.

\bibitem[\protect\citeauthoryear{Farajtabar \bgroup \em et al.\egroup }{2020}]{farajtabar2020orthogonal}
Mehrdad Farajtabar, Navid Azizan, Alex Mott, and Ang Li.
\newblock Orthogonal gradient descent for continual learning.
\newblock In {\em International Conference on Artificial Intelligence and Statistics}, pages 3762--3773. PMLR, 2020.

\bibitem[\protect\citeauthoryear{He and Jaeger}{2018}]{he2018overcoming}
Xu~He and Herbert Jaeger.
\newblock Overcoming catastrophic interference using conceptor-aided backpropagation.
\newblock In {\em International Conference on Learning Representations}, 2018.

\bibitem[\protect\citeauthoryear{He \bgroup \em et al.\egroup }{2016}]{he2016deep}
Kaiming He, Xiangyu Zhang, Shaoqing Ren, and Jian Sun.
\newblock Deep residual learning for image recognition.
\newblock In {\em Proceedings of the IEEE conference on computer vision and pattern recognition}, pages 770--778, 2016.

\bibitem[\protect\citeauthoryear{He \bgroup \em et al.\egroup }{2022}]{DBLP:conf/iclr/HeZMBN22}
Junxian He, Chunting Zhou, Xuezhe Ma, Taylor Berg{-}Kirkpatrick, and Graham Neubig.
\newblock Towards a unified view of parameter-efficient transfer learning.
\newblock In {\em The Tenth International Conference on Learning Representations, {ICLR} 2022, Virtual Event, April 25-29, 2022}. OpenReview.net, 2022.

\bibitem[\protect\citeauthoryear{Houlsby \bgroup \em et al.\egroup }{2019}]{adapter}
Neil Houlsby, Andrei Giurgiu, Stanislaw Jastrzebski, Bruna Morrone, Quentin De~Laroussilhe, Andrea Gesmundo, Mona Attariyan, and Sylvain Gelly.
\newblock Parameter-efficient transfer learning for nlp.
\newblock In {\em International Conference on Machine Learning}, pages 2790--2799. PMLR, 2019.

\bibitem[\protect\citeauthoryear{Hu \bgroup \em et al.\egroup }{2021}]{lora}
Edward~J Hu, Yelong Shen, Phillip Wallis, Zeyuan Allen-Zhu, Yuanzhi Li, Shean Wang, Lu~Wang, and Weizhu Chen.
\newblock Lora: Low-rank adaptation of large language models.
\newblock {\em arXiv preprint arXiv:2106.09685}, 2021.

\bibitem[\protect\citeauthoryear{Karimi~Mahabadi \bgroup \em et al.\egroup }{2021}]{karimi2021compacter}
Rabeeh Karimi~Mahabadi, James Henderson, and Sebastian Ruder.
\newblock Compacter: Efficient low-rank hypercomplex adapter layers.
\newblock {\em Advances in Neural Information Processing Systems}, 34:1022--1035, 2021.

\bibitem[\protect\citeauthoryear{Kemker and Kanan}{2018}]{DBLP:conf/iclr/KemkerK18}
Ronald Kemker and Christopher Kanan.
\newblock Fearnet: Brain-inspired model for incremental learning.
\newblock In {\em 6th International Conference on Learning Representations, {ICLR} 2018, Vancouver, BC, Canada, April 30 - May 3, 2018, Conference Track Proceedings}. OpenReview.net, 2018.

\bibitem[\protect\citeauthoryear{Kirkpatrick \bgroup \em et al.\egroup }{2017}]{regular_kirk}
James Kirkpatrick, Razvan Pascanu, Neil Rabinowitz, Joel Veness, Guillaume Desjardins, Andrei~A Rusu, Kieran Milan, John Quan, Tiago Ramalho, Agnieszka Grabska-Barwinska, et~al.
\newblock Overcoming catastrophic forgetting in neural networks.
\newblock {\em Proceedings of the national academy of sciences}, 114(13):3521--3526, 2017.

\bibitem[\protect\citeauthoryear{Lopez-Paz and Ranzato}{2017}]{replay_lopez}
David Lopez-Paz and Marc'Aurelio Ranzato.
\newblock Gradient episodic memory for continual learning.
\newblock {\em Advances in neural information processing systems}, 30, 2017.

\bibitem[\protect\citeauthoryear{Maas \bgroup \em et al.\egroup }{2011}]{maas2011learning}
Andrew Maas, Raymond~E Daly, Peter~T Pham, Dan Huang, Andrew~Y Ng, and Christopher Potts.
\newblock Learning word vectors for sentiment analysis.
\newblock In {\em Proceedings of the 49th annual meeting of the association for computational linguistics: Human language technologies}, pages 142--150, 2011.

\bibitem[\protect\citeauthoryear{Mallya and Lazebnik}{2018}]{mallya2018packnet}
Arun Mallya and Svetlana Lazebnik.
\newblock Packnet: Adding multiple tasks to a single network by iterative pruning.
\newblock In {\em Proceedings of the IEEE conference on Computer Vision and Pattern Recognition}, pages 7765--7773, 2018.

\bibitem[\protect\citeauthoryear{Mallya \bgroup \em et al.\egroup }{2018}]{mallya2018piggyback}
Arun Mallya, Dillon Davis, and Svetlana Lazebnik.
\newblock Piggyback: Adapting a single network to multiple tasks by learning to mask weights.
\newblock In {\em Proceedings of the European conference on computer vision (ECCV)}, pages 67--82, 2018.

\bibitem[\protect\citeauthoryear{Qin and Joty}{2021}]{qin2021lfpt5}
Chengwei Qin and Shafiq Joty.
\newblock Lfpt5: A unified framework for lifelong few-shot language learning based on prompt tuning of t5.
\newblock {\em arXiv preprint arXiv:2110.07298}, 2021.

\bibitem[\protect\citeauthoryear{Raffel \bgroup \em et al.\egroup }{2020}]{raffel2020exploring}
Colin Raffel, Noam Shazeer, Adam Roberts, Katherine Lee, Sharan Narang, Michael Matena, Yanqi Zhou, Wei Li, and Peter~J Liu.
\newblock Exploring the limits of transfer learning with a unified text-to-text transformer.
\newblock {\em The Journal of Machine Learning Research}, 21(1):5485--5551, 2020.

\bibitem[\protect\citeauthoryear{Razdaibiedina \bgroup \em et al.\egroup }{2023}]{razdaibiedina2023progressive}
Anastasia Razdaibiedina, Yuning Mao, Rui Hou, Madian Khabsa, Mike Lewis, and Amjad Almahairi.
\newblock Progressive prompts: Continual learning for language models.
\newblock {\em arXiv preprint arXiv:2301.12314}, 2023.

\bibitem[\protect\citeauthoryear{Rebuffi \bgroup \em et al.\egroup }{2017}]{rebuffi2017icarl}
Sylvestre-Alvise Rebuffi, Alexander Kolesnikov, Georg Sperl, and Christoph~H Lampert.
\newblock icarl: Incremental classifier and representation learning.
\newblock In {\em Proceedings of the IEEE conference on Computer Vision and Pattern Recognition}, pages 2001--2010, 2017.

\bibitem[\protect\citeauthoryear{Rusu \bgroup \em et al.\egroup }{2016}]{rusu2016progressive}
Andrei~A Rusu, Neil~C Rabinowitz, Guillaume Desjardins, Hubert Soyer, James Kirkpatrick, Koray Kavukcuoglu, Razvan Pascanu, and Raia Hadsell.
\newblock Progressive neural networks.
\newblock {\em arXiv preprint arXiv:1606.04671}, 2016.

\bibitem[\protect\citeauthoryear{Shin \bgroup \em et al.\egroup }{2017}]{shin2017continual}
Hanul Shin, Jung~Kwon Lee, Jaehong Kim, and Jiwon Kim.
\newblock Continual learning with deep generative replay.
\newblock {\em Advances in neural information processing systems}, 30, 2017.

\bibitem[\protect\citeauthoryear{Touvron \bgroup \em et al.\egroup }{2023}]{touvron2023llama}
Hugo Touvron, Louis Martin, Kevin Stone, Peter Albert, Amjad Almahairi, Yasmine Babaei, Nikolay Bashlykov, Soumya Batra, Prajjwal Bhargava, Shruti Bhosale, et~al.
\newblock Llama 2: Open foundation and fine-tuned chat models.
\newblock {\em arXiv preprint arXiv:2307.09288}, 2023.

\bibitem[\protect\citeauthoryear{Wang \bgroup \em et al.\egroup }{2018}]{wang2018glue}
Alex Wang, Amanpreet Singh, Julian Michael, Felix Hill, Omer Levy, and Samuel~R Bowman.
\newblock Glue: A multi-task benchmark and analysis platform for natural language understanding.
\newblock {\em arXiv preprint arXiv:1804.07461}, 2018.

\bibitem[\protect\citeauthoryear{Wang \bgroup \em et al.\egroup }{2019}]{wang2019superglue}
Alex Wang, Yada Pruksachatkun, Nikita Nangia, Amanpreet Singh, Julian Michael, Felix Hill, Omer Levy, and Samuel Bowman.
\newblock Superglue: A stickier benchmark for general-purpose language understanding systems.
\newblock {\em Advances in neural information processing systems}, 32, 2019.

\bibitem[\protect\citeauthoryear{Wang \bgroup \em et al.\egroup }{2022}]{wang2022learning}
Zifeng Wang, Zizhao Zhang, Chen-Yu Lee, Han Zhang, Ruoxi Sun, Xiaoqi Ren, Guolong Su, Vincent Perot, Jennifer Dy, and Tomas Pfister.
\newblock Learning to prompt for continual learning.
\newblock In {\em Proceedings of the IEEE/CVF Conference on Computer Vision and Pattern Recognition}, pages 139--149, 2022.

\bibitem[\protect\citeauthoryear{Wang \bgroup \em et al.\egroup }{2023a}]{olora}
Xiao Wang, Tianze Chen, Qiming Ge, Han Xia, Rong Bao, Rui Zheng, Qi~Zhang, Tao Gui, and Xuanjing Huang.
\newblock Orthogonal subspace learning for language model continual learning.
\newblock {\em arXiv preprint arXiv:2310.14152}, 2023.

\bibitem[\protect\citeauthoryear{Wang \bgroup \em et al.\egroup }{2023b}]{trace}
Xiao Wang, Yuansen Zhang, Tianze Chen, Songyang Gao, Senjie Jin, Xianjun Yang, Zhiheng Xi, Rui Zheng, Yicheng Zou, Tao Gui, et~al.
\newblock Trace: A comprehensive benchmark for continual learning in large language models.
\newblock {\em arXiv preprint arXiv:2310.06762}, 2023.

\bibitem[\protect\citeauthoryear{Zenke \bgroup \em et al.\egroup }{2017}]{regular_zenke}
Friedemann Zenke, Ben Poole, and Surya Ganguli.
\newblock Continual learning through synaptic intelligence.
\newblock In {\em International conference on machine learning}, pages 3987--3995. PMLR, 2017.

\bibitem[\protect\citeauthoryear{Zhang \bgroup \em et al.\egroup }{2015}]{zhang2015character}
Xiang Zhang, Junbo Zhao, and Yann LeCun.
\newblock Character-level convolutional networks for text classification.
\newblock {\em Advances in neural information processing systems}, 28, 2015.

\bibitem[\protect\citeauthoryear{Zhu \bgroup \em et al.\egroup }{2021}]{zhu2021counter}
Yaoming Zhu, Jiangtao Feng, Chengqi Zhao, Mingxuan Wang, and Lei Li.
\newblock Counter-interference adapter for multilingual machine translation.
\newblock In {\em Findings of the Association for Computational Linguistics: EMNLP 2021}, pages 2812--2823, 2021.

\end{thebibliography}

%%%%%%%%%%%%%%%%%%%%%%%%%%%%%%%%%%%%%%%%%%%%%%%%%%%%%%%%%%%%
\clearpage

\section{Appendix}

\subsection{Details of The Training Tasks}

In this section, we report the details of the tasks and datasets in our experiments. 

Table \ref{Table 3} shows the details of all 15 datasets in the standard CL benchmark and the Large number of tasks.  Overall, we used 5 datasets from CL benchmark [Zhang et al., 2015], 4 datasets from GLUE [Wang et al., 2018] and 6 datasets from SuperGLUE [Wang et al., 2019] benchmarks, following [Razdaibiedina et al., 2023].

Following [Wang et al., 2023a], We report task orders used for our experiments across T5-Large and LLaMA2-7B models in Table \ref{Table 4}. Furthermore, we use the instruction tuning method, so we have corresponding instructions for each task. Table \ref{Table 5} shows prompts for different tasks. NLI denotes natural language inference, including MNLI, RTE and CB. SC denotes sentiment analysis, including Amazon, Yelp, SST-2 and IMDB. TC denotes topic classification, including AG News, Dbpedia and Yahoo.

\subsection{Additional computation of AM-LoRA}

Similar to O-LoRA, AM-LoRA will increase the number of LoRA's parameters by one for each new task trained. In addition, AM-LoRA adds an additional Attentional Selector of size 4096*1. For example, in LLaMA2-7B, the AM-LoRA method adds an additional 4096*2*32 number of trainable senators, which is 0.00003 of the total number of senators, and this amount can be considered as minimal.

\subsection{Implementation Details}

For all orders of task streams, We trained the models with one epoch, a constant learning rate of 1e-4, a batch size of 1 per GPU in LLaMA2-7B, a batch size of 8 per GPU in T5-Large, a dropout rate of 0.1, and a weight decay rate of 0.

The values of $\lambda$ are different among order 1 to 6. For order 1, order 2, order 3
in T5-Large and LLaMA2-7B, we set $\lambda$ = 1e-5, 1e-5, 1e-5, 1e-5.  For every task in order 4(MNLI, CB, WiC, COPA, QQP, BoolQA, RTE, IMDB, Yelp, Amazon, SST-2, DBpedia, Agnews, MultiRC, Yahoo), we set $\lambda$ = 0, 0.01, 0.01, 0.01, 0.01, 0.01, 0.3, 0.001, 0.001, 0.001, 0.001, 0.001, 0.001, 0.001, 0.001. For order 5(MultiRC, BoolQA, WiC, MNLI, CB, COPA, QQP, RTE, IMDB, SST-2, DBpedia, Agnews, Yelp, Amazon, Yahoo), we set $\lambda$ = 0.0001, 0.0001, 0.001, 0.001, 0.001, 0.001, 0.001, 0.001, 0.001, 0.001, 0.01, 0.01, 0.01, 0.01, 0.01. For order 6(Yelp, Amazon, MNLI, CB, COPA, QQP, RTE, IMDB, SST-2, DBpedia, Agnews, Yahoo, MultiRC, BoolQA, WiC), we set $\lambda$ = 0.001, 0.001, 0.001, 0.001, 0.001, 0.001, 0.001, 0.001, 0.001, 0.001, 0.001, 0.001, 0.001, 0.001, 0.001 respectively. 

\subsection{Supplementary Explanation}

We analyzed why orthogonal constraints cannot solve catastrophic forgetting from the perspective of parameter space in the introduction section. In this section, we will mathematically prove that the method of limiting the orthogonality between various LoRA matrices cannot solve catastrophic forgetting.

Considering the case of sequential learning of two tasks, we assume that the model after learning the first task is represented as \(f(Ax)\). If the second task is learned on the basis of \(f(Ax)\), the model can be expressed as \(f((A+B)x)\). Where \(A\) is the weight of the first task, \(B \) is the weight of the second task. According to the distributive law, it is easy to derive the following equation \(f((A+B)x) = f(Ax + Bx)\). Even if there is \(A^T B = 0\), it does not mean that \(Bx\) is equal to the zero matrix, because \(B\) is not necessarily a zero matrix. Therefore, the condition of \(A^T B = 0\) is not enough to guarantee \(f(Ax) = f((A+B)x)\) without additional constraints. Because of the existence and uncertainty of \(Bx\), the form of \(f(x)\) may cause the two functions to produce different outputs. In addition, we also give more specific examples in Appendix A.1. Furthermore, we conducted an empirical study in our experiment. In summary, learning the parameters of a new task in an orthogonal space is insufficient to solve the catastrophic forgetting problem in LLM scenarios. The following is explained through specific examples.

Assume that in the classification task, the parameter matrices of tasks A and B are orthogonal to each other. We need to show that the predicted labels of the model after training tasks A and B sequentially will be biased compared to the predicted labels of training only task A.

\begin{table*}
    \centering
    \begin{tabularx}{\linewidth}{@{\hspace{0.05em}}p{2cm}@{\hspace{0.05em}}p{3cm}@{\hspace{0.05em}}p{3.5cm}@{\hspace{0.05em}}p{3.5cm}@{\hspace{0.05em}}X}
        \toprule
        \textbf{Dataset} & \textbf{Benchmark} & \textbf{Task} & \textbf{Domain} & \textbf{Metric}\\
        \midrule
        Yelp & CL Benchmark & Sentiment Analysis & Yelp Reviews & Accuracy\\
        Amazon & CL Benchmark & Sentiment Analysis & Amazon Reviews & Accuracy\\
        DBpedia & CL Benchmark & Topic Classification & Wikipedia & Accuracy\\
        Yahoo & CL Benchmark & Topic Classification & Yahoo Q\&A & Accuracy\\
        AG News & CL Benchmark & Topic Classification & News & Accuracy\\
        MNLI & GLUE & NLI & Various & Accuracy\\
        QQP & GLUE & Paragraph Detection & Quora & Accuracy\\
        RTE & GLUE & NLI & News, Wikipedia & Accuracy\\
        SST-2 & GLUE & Sentiment Analysis & Movie Reviews & Accuracy\\
        WiC & SuperGLUE & \makecell[tl]{Word Sense \\ Disambiguation} & Lexical Databases & Accuracy\\
        CB & SuperGLUE & NLI & Various & Accuracy\\
        COPA & SuperGLUE & QA & Blogs,Encyclopedia & Accuracy\\
        BoolQA & SuperGLUE & Boolean QA & Wikipedia & Accuracy\\
        MultiRC & SuperGLUE & QA & Various & Accuracy\\
        IMDB & SuperGLUE & Sentiment Analysis & Movie Reviews & Accuracy\\
        \bottomrule
    \end{tabularx}
    \caption{The details of 15 datasets used in our CL experiments. NLI denotes natural language inference, QA denotes questions and answers task. First five tasks correspond to the standard CL benchmark, all other tasks are used in long-sequence experiments.}
    \label{Table 3}
\end{table*}

\begin{table*}
    \centering
    \begin{tabularx}{\linewidth}{p{1.5cm}p{4.5cm}X}
        \toprule
        \textbf{Order} & \textbf{Model} & \textbf{Task Sequence}\\
        \midrule
        1 & T5-Large, LLaMA2-7B & dbpedia $\rightarrow$ amazon $\rightarrow$ yahoo $\rightarrow$ ag\\
        2 & T5-Large, LLaMA2-7B & dbpedia $\rightarrow$ amazon $\rightarrow$ ag $\rightarrow$ yahoo\\
        3 & T5-Large, LLaMA2-7B & yahoo $\rightarrow$ amazon $\rightarrow$ ag $\rightarrow$ dbpedia\\
        \multirow{2}{*}{4} & \multirow{2}{*}{T5-Large} & mnli $\rightarrow$ cb $\rightarrow$ wic $\rightarrow$ copa $\rightarrow$ qqp $\rightarrow$ boolqa $\rightarrow$ rte $\rightarrow$ imdb $\rightarrow$ yelp $\rightarrow$ amazon $\rightarrow$ sst-2 $\rightarrow$ dbpedia $\rightarrow$ ag $\rightarrow$ multirc $\rightarrow$ yahoo\\
        \multirow{2}{*}{5} & \multirow{2}{*}{T5-Large} & multirc $\rightarrow$ boolqa $\rightarrow$ wic $\rightarrow$ mnli $\rightarrow$ cb $\rightarrow$ copa $\rightarrow$ qqp $\rightarrow$ rte $\rightarrow$ imdb $\rightarrow$ sst-2 $\rightarrow$ dbpedia $\rightarrow$ ag $\rightarrow$ yelp $\rightarrow$ amazon $\rightarrow$ yahoo\\
        \multirow{2}{*}{6} & \multirow{2}{*}{T5-Large} & yelp $\rightarrow$ amazon $\rightarrow$ mnli $\rightarrow$ cb $\rightarrow$ copa $\rightarrow$ qqp $\rightarrow$ rte $\rightarrow$ imdb $\rightarrow$ sst-2 $\rightarrow$ dbpedia $\rightarrow$ ag $\rightarrow$ yahoo $\rightarrow$ multirc $\rightarrow$ boolqa $\rightarrow$ wic\\
        \bottomrule
    \end{tabularx}
    \caption{Six different orders of task sequences used for continual learning experiments. Orders 1-3 correspond to the standard CL becnhmark adopted by prior works. Orders 4-6 are long-sequence orders spanning 15 tasks, following [Razdaibiedina et al., 2023].}
    \label{Table 4}
\end{table*}

\begin{table*}
    \centering
    \begin{tabularx}{\linewidth}{lX}
        \toprule
        \textbf{Task} & \textbf{Prompts}\\
        \midrule
        NLI & What is the logical relationship between the “sentence 1” and the “sentence 2”? Choose one from the option.\\
        QQP & Whether the “first sentence" and the “second sentence" have the same meaning? Choose one from the option.\\
        SC & What is the sentiment of the following paragraph? Choose one from the option.\\
        TC & What is the topic of the following paragraph? Choose one from the option.\\
        BoolQA & According to the following passage, is the question true or false? Choose one from the option.\\
        MultiRC & According to the following passage and question, is the candidate answer true or false? Choose one from the option.\\
        WiC & Given a word and two sentences, whether the word is used with the same sense in both sentences? Choose one from the option.\\
        \bottomrule
    \end{tabularx}
    \caption{Instructions for different tasks.}
    \label{Table 5}
\end{table*}

\subsubsection{One Dimensional Space}
Consider the one-dimensional case, where \( f(x) = \sin x \), \( A = (1, 0) \), \( B = (0, -1) \), satisfying \( A^TB = 0 \). When \( x = \left(\frac{\pi}{2}, \pi\right) \), \( \sin(Ax) = 1 \), \( \sin((A+B)x) = -1 \). It can be seen that even if matrix B is orthogonal to A, the predicted label after adding matrix B will produce deviations.

\subsubsection{Two-dimensional space}
Let $A$ be the matrix given by

\[
A = \begin{bmatrix}
    1 & 0 \\
    0 & 0 \\
\end{bmatrix},
\tag{7}
\]
and let $B$ be the matrix
\[
B = \begin{bmatrix}
    0 & 0 \\
    0 & 1 \\
\end{bmatrix}.
\tag{8}
\]
These matrices satisfy $A^TB = 0$. Now, consider the vector $x$ given by
\[
x = \begin{bmatrix}
    1 \\
    -1 \\
\end{bmatrix}.
\tag{9}
\]
We find that $Ax$ is
\[
Ax = \begin{bmatrix}
    1 \\
    0 \\
\end{bmatrix},
\tag{10}
\]
and $(A+B)x$ is
\[
(A+B)x = \begin{bmatrix}
    1 \\
    -1 \\
\end{bmatrix}.
\tag{11}
\]
Define a function $f(x)$ as
\[
f(x) = \begin{bmatrix}
    x_1 + x_2 \\
    -x_2 \\
\end{bmatrix}.
\tag{12}
\]
This function can also be expressed as the product of matrices:
\[
f(x) = \begin{bmatrix}
    1 & 1 \\
    0 & -1 \\
\end{bmatrix} \begin{bmatrix}
    x_1 \\
    x_2 \\
\end{bmatrix}.
\tag{13}
\]
Now, let's evaluate $f(Ax)$:
\[
f(Ax) = \begin{bmatrix}
    1 \\
    0 \\
\end{bmatrix},
\tag{14}
\]
and $f((A+B)x)$ is
\[
f((A+B)x) = \begin{bmatrix}
    0 \\
    1 \\
\end{bmatrix}.
\tag{15}
\]
So in the two-dimensional case, there is also $f(Ax) \neq f((A+B)x)$.

\subsubsection{N-dimensional space}

Finally, it is extended to n dimensions, let \(A\) be the matrix given by

\[ A = \begin{bmatrix}
	1 & 0 & \cdots & 0 \\
	0 & 0 & \cdots & 0 \\
	\vdots & \vdots & \ddots & \vdots \\
	0 & 0 & \cdots & 0 \\
\end{bmatrix}_{n\times n}, 
\tag{16}
\]

and let \(B\) be the matrix given by

\[ B = \begin{bmatrix}
	0 & \cdots & 0 & 0 \\
	\vdots & \ddots & \vdots & \vdots \\
	0 & \cdots & 0 & 0 \\
	0 & \cdots & 0 & 1 \\
\end{bmatrix}_{n\times n}. 
\tag{17}
\]

If \(A^TB = 0\), and we define \(x\) as

\[ x = \begin{bmatrix}
	1 \\
	0 \\
	\vdots \\
	0 \\
	-1 \\
\end{bmatrix}_{n\times 1}, 
\tag{18}
\]

then we have \(Ax = \begin{bmatrix}
	1 \\
	0 \\
	\vdots \\
	0 \\
\end{bmatrix}_{n\times 1}\) and \((A+B)x = \begin{bmatrix}
	1 \\
	0 \\
	\vdots \\
	0 \\
	-1 \\
\end{bmatrix}_{n\times 1}\). \\[5pt]

Now, let's define a function \(f(x)\) as

\[ f(x) = \begin{bmatrix}
	x_1 + x_n \\
	0 \\
	\vdots \\
	0 \\
	-x_n \\
\end{bmatrix} = \begin{bmatrix}
	1 & 0 & \cdots & 0 & 1 \\
	0 & 0 & \cdots & 0 & 0 \\
	\vdots & \vdots & \ddots & \vdots & \vdots \\
	\vdots & 0 & \cdots & 0 & 0 \\
	0 & \cdots & \cdots & 0 & -1 \\
\end{bmatrix} \begin{bmatrix}
	x_1 \\
	\vdots \\
	x_n \\
\end{bmatrix}. 
\tag{19}
\]

Therefore, we get \(f(Ax) = \begin{bmatrix}
	1 \\
	\vdots \\
	0 \\
\end{bmatrix}_{n\times 1}\) and \(f((A+B)x) = \begin{bmatrix}
	0 \\
	\vdots \\
	1 \\
\end{bmatrix}_{n\times 1}\). \\[5pt]

So in the case of n dimensions, there is also \(f(Ax)\neq f((A+B)x)\).

We can also find many similar examples. From the above, we can conclude that adding a new parameter matrix that is orthogonal to the previous parameter matrix will bias the prediction. In other words, using multiple mutually orthogonal LoRA matrices to learn sequence tasks in a continuous learning scenario cannot avoid catastrophic forgetting.

\end{document}